\documentclass{article}

\usepackage[preprint]{neurips_data_2024}
\usepackage[utf8]{inputenc} % allow utf-8 input
\usepackage[T1]{fontenc}    % use 8-bit T1 fonts
\usepackage[hidelinks]{hyperref}       % hyperlinks
\usepackage{nicefrac}       % compact symbols for 1/2, etc.
\usepackage{microtype}      % microtypography
\usepackage{rotating}
\usepackage{graphicx}
\usepackage{nameref}
\usepackage{subcaption}
\usepackage{amsmath,amsthm,trimclip}
\usepackage{amssymb}
\usepackage{amsfonts}
\usepackage{footnote}
\usepackage{url}
\usepackage{lipsum}
\usepackage[inline]{enumitem}
\usepackage[acronym]{glossaries}
\usepackage{bbm}
\usepackage{adjustbox}
\usepackage{algorithm} 
\usepackage{algpseudocode}
\usepackage{booktabs}
\usepackage{mdframed}
\usepackage{color, colortbl}
\usepackage[table]{xcolor}
\usepackage{placeins}
\usepackage{capt-of}
\usepackage[frozencache,cachedir=.]{minted}
\usepackage{float}
\usepackage{newfloat}

\setminted{fontsize=\footnotesize}
\setminted{frame=leftline,framerule=1pt}

\definecolor{gemma}{HTML}{1B98E0}
\definecolor{llama}{HTML}{76C7C0}
\definecolor{mistral}{HTML}{FFD79D}
\definecolor{commandr}{HTML}{02475E}

\newtheoremstyle{postulate}{1em}{1em}{}{}{\bfseries}{\normalfont\textit{.}}{.5em}{}
\theoremstyle{postulate}

\floatstyle{plain}
\newfloat{snippet}{H}{lop}
\floatname{snippet}{Code}

\author{%
Eduardo Pignatelli \\
University College London \\
\texttt{e.pignatelli@ucl.ac.uk}
\And
Jarek Liesen \\
BCCN Berlin \\
\texttt{jarek@bccn-berlin.de}
\And
Robert Tjarko Lange \\
Technical University Berlin \\
\texttt{robert.t.lange@tu-berlin.de}
\And
Chris Lu \\
University of Oxford \\
\texttt{christopher.lu@eng.ox.ac.uk}
\And
Pablo Samuel Castro \\
Google DeepMind \\
Mila, Université de Montréal \\
\texttt{psc@google.com}
\And
Laura Toni \\
University College London \\
\texttt{l.toni@ucl.ac.uk} \\
}

\begin{document}

\title{NAVIX: Scaling MiniGrid Environments with JAX}
\maketitle

% - Acronyms
\newacronym{rl}{RL}{Reinforcement Learning}
\newacronym{drl}{Deep RL}{Deep Reinforcement Learning}
\newacronym[longplural=Markov Decision Processes]{mdp}{MDP}{Markov Decision Process}
\newacronym[longplural=Markov Processes]{mp}{MP}{Markov Process}
\newacronym{pomdp}{POMDP}{Partially-observable MDP}
\newacronym[longplural=Markov Reward Processes]{mrp}{MRP}{Markov Reward Process}
\newacronym[longplural=Random Processes]{rp}{RP}{Random Process}
\newacronym{ca}{CA}{Credit Assignment}
\newacronym{cap}{CAP}{Credit Assignment Problem}
\newacronym{hi}{HI}{Hindsight Information}
\newacronym{gpi}{GPI}{Generalised Policy Iteration}
\newacronym{mpi}{MPI}{Modified Policy Iteration}
\newacronym{pe}{PE}{Policy Evaluation}
\newacronym{pi}{PI}{Policy Improvement}
\newacronym{lstm}{LSTM}{Long-Short-Term-Memory}
\newacronym{td}{TD}{Temporal Difference}
\newacronym{kl}{KL}{Kullback-Leibler}
\newacronym{dnn}{DNN}{Deep Neural Network}
\newacronym{llm}{LLM}{Large Language Model}
\newacronym{vlm}{LLM}{Vision Language Model}
\newacronym{calm}{CALM}{Credit Assignment with Language Models}
\newacronym{hrl}{HRL}{Hierarchical Reinforcement Learning}
\newacronym{xla}{XLA}{Accelerated Linear Algebra}
\newacronym{ecsm}{ECSM}{Entity-Component-System Model}
% ---

% ---------------------------------------------------------------------------
% - Abstract   --------------------------------------------------------------
% ---------------------------------------------------------------------------
\begin{abstract}
    As \gls{drl} research moves towards solving large-scale worlds, efficient environment simulations become crucial for rapid experimentation.
    However, most existing environments struggle to scale to high throughput, setting back meaningful progress.
    Interactions are typically computed on the CPU, limiting training speed and throughput, due to slower computation and communication overhead when distributing the task across multiple machines.
    Ultimately, \gls{drl} training is CPU-bound, and developing batched, fast, and scalable environments has become a frontier for progress.
    % Recently, a set of GPU-based environments has sparked raising interest, proposing a JAX-based, batched re-implementations of common \gls{rl} environments that significantly increase the throughput of canonical \gls{drl} algorithms, and enabling large scale parallelism.
    % These environments allow training thousands of agents simultaneously on a single accelerator, vastly outperforming traditional CPU-based environments.
    %
    Among the most used \gls{rl} environments, MiniGrid is at the foundation of several studies on exploration, curriculum learning, representation learning, diversity, meta-learning, credit assignment, and language-conditioned \gls{rl}, and still suffers from the limitations described above.
    In this work, we introduce NAVIX\footnote{https://github.com/epignatelli/navix}, a re-implementation of MiniGrid in JAX.
    NAVIX achieves over $200\,000\times$ speed improvements in batch mode, supporting up to 2048 agents in parallel on a single Nvidia A100 80 GB.
    This reduces experiment times from one week to 15 minutes, promoting faster design iterations and more scalable \gls{rl} model development.
\end{abstract}

\begin{figure}[h]
    \centering
    \begin{subfigure}{.19\textwidth}
        \includegraphics[width=.95\linewidth]{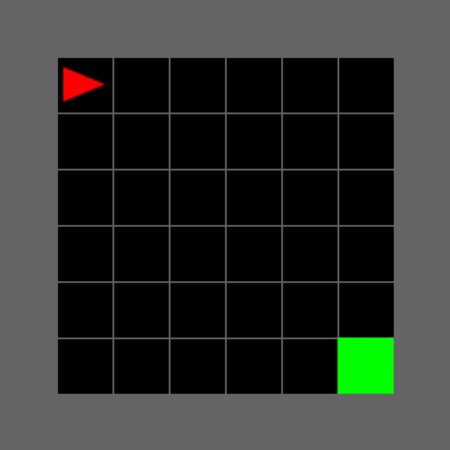}
        \caption{128.98$\times$}
    \end{subfigure}
    \hfill
    \begin{subfigure}{.19\textwidth}
        \includegraphics[width=.95\linewidth]{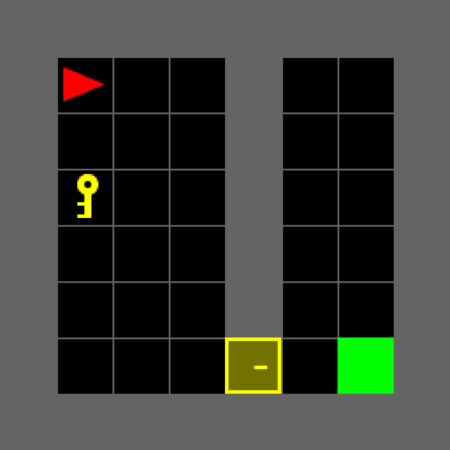}
        \caption{26.47$\times$}
    \end{subfigure}
    \hfill
    \begin{subfigure}{.19\textwidth}
        \includegraphics[width=.95\linewidth]{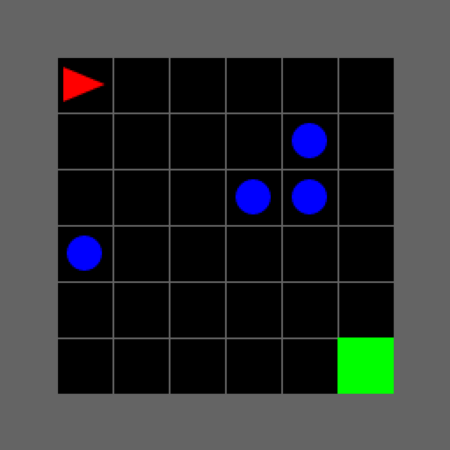}
        \caption{41.17$\times$}
    \end{subfigure}
    \hfill
    \begin{subfigure}{.19\textwidth}
        \includegraphics[width=.95\linewidth]{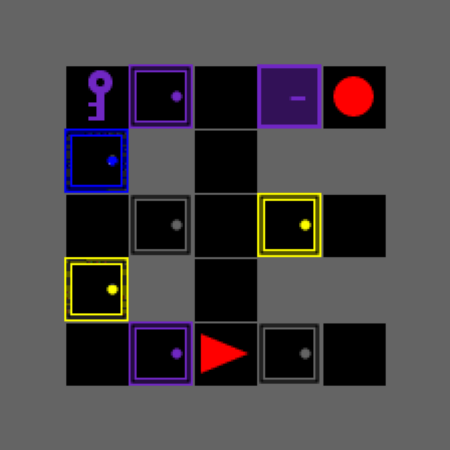}
        \caption{19.72$\times$}
    \end{subfigure}
    \hfill
    \begin{subfigure}{.19\textwidth}
        \includegraphics[width=.95\linewidth]{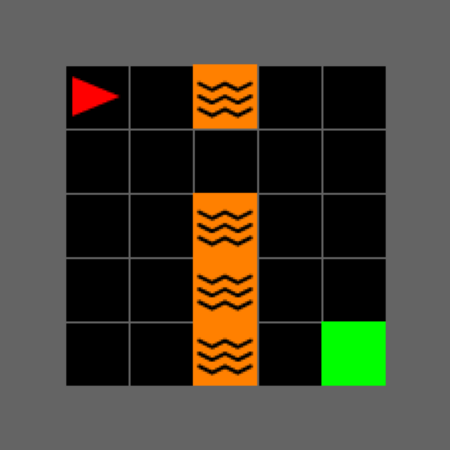}
        \caption{45.71$\times$}
    \end{subfigure}
    \caption{
        Speedups for five of the NAVIX environments with respect to their MiniGrid equivalent, using the protocol in Section~\ref{sec:experiments:speed}.
        (a) \texttt{Empty-8x8-v0}, (b) \texttt{DoorKey-8x8-v0}, (c) \texttt{Dynamic-Obstacles-8x8-v0}, (d) \texttt{KeyCorridorS3R3-v0}, (e) \texttt{LavaGapS7-v0}.
    }
    \label{fig:cover}
\end{figure}
% ---

% ---------------------------------------------------------------------------
% - Introduction  -----------------------------------------------------------
% ---------------------------------------------------------------------------
\section{Introduction}
\label{sec:intro}
% Context (RL is data hungry)
\glsfirst{drl} is notoriously sample inefficient \citep{kaiser2019model,wang2021alchemy,johnson2016malmo,kuttler2020nethack}.
Depending on the complexity of the environment dynamics, the observation space, and the action space, agents often require between $10^7$ to $10^9$ interactions or even more for training up to a good enough policy.
Therefore, as \gls{drl} moves towards tackling more complex environments, leveraging efficient environment implementations is an essential ingredient of rapid experimentation and fast design iterations.

% Limitations of current CPU-based envs
However, while the efficiency and scalability of solutions for \textit{agents} have improved massively in recent years \citep{schulman2017proximal,espeholt2018impala,kapturowski2018recurrent}, especially due to the scalability of the current deep learning frameworks \citep{abadi2016tensorflow,paszke2019pytorch,ansel2024pytorch,jax2018github,xla2020xla}, environments have not kept pace.
They are mostly based on CPU, cannot adapt to different types of devices, and scaling often requires complex distributed systems, introducing design complexity and communication overhead.
Overall, deep \gls{rl} experiments are CPU-bound, limiting both speed and throughput of \gls{rl} training.

% Current JAX-based envs
Recently, a set of GPU-based environments \citep{brax2021github,gymnax2022github,weng2022envpool,koyamada2023pgx,flair2023jaxmarl,nikulin2023xlandMiniGrid,matthews2024craftax,bonnet2024jumanji, lu2023structured, liesen2024discovering} and frameworks \citep{lu2022discovered,liesen2024rejax,toledo2024stoix,nishimori2022jaxcorl,jiang2023minimax} has sparked raising interest, proposing JAX-based, batched implementations of common \gls{rl} environments that can significantly increase the speed and throughput of canonical \gls{drl} algorithms.
This enables large-scale parallelism, allowing the training of thousands of agents in parallel on a single accelerator, significantly outperforming traditional CPU-based environments, and fostering meta-\gls{rl} applications.

% Minigrid
In this work, we build on this trend and focus on the MiniGrid suite of environments \citep{chevalier2024minigrid}, due to its central role in the \gls{drl} literature.
MiniGrid is fundamental to many studies.
For instance, \citet{zhang2020bebold,zha2021rank,mavor2022stay} used it to test new exploration strategies; \citet{jiang2021prioritized} for curriculum learning; \citet{zhao2021consciousness} for planning; \citet{paischer2022history} for representation learning, \citet{flet2021adversarially,guan2022leveraging} for diversity. \citet{parisi2021interesting} employed MiniGrid to design meta and transfer learning strategies, and \citet{mu2022improving} to study language grounding.

However, despite its ubiquity in the \gls{drl} literature, MiniGrid faces the limitations of CPU-bound environments.
We bridge this gap and propose NAVIX, a reimplementation of Minigrid in JAX that leverages JAX's intermediate language representation to migrate the computation to different accelerators, such as GPUs, and TPUs.

% Results
Our results show that NAVIX is over $10\times$ faster than the original Minigrid implementation, in common \gls{drl} settings (see Section~\ref{sec:experiments:speed}), and increases the throughput by over $10^6\times$, turning 1-week experiments into 15 minutes ones.
We show the scaling ability of NAVIX by training over 2048 PPO agents in parallel (see Section~\ref{sec:experiments:throughput}), each using their own subset of environments, all on a single Nvidia A100 80 GB.

The main contributions of this work are the following:
\begin{enumerate}
    \item A fully JAX-based implementation of environment configurations that reproduces exactly the original Minigrid \glspl{mdp} and \glspl{pomdp}.
    \item A description of the design philosophy, the design pattern and principles, the organisation, and the components of NAVIX, which, together with the online documentation\footnote{https://epignatelli/navix}, form an instruction manual to use and extend NAVIX.
    \item A set of \gls{rl} algorithm baselines for all environments in Section~\ref{sec:experiments:baselines}.
    % \item A scoreboard,\footnote{Available at \url{https://wandb.ai/edu-pignatelli/navix-baselines/}.} storing scores for common algorithms that new method can use for baseline comparisons, which otherwise can be prone to errors and manipulations. This offers a clearer, broader picture of the current space of algorithms.
\end{enumerate}
% ---------------------------------------------------------------------------

% ---------------------------------------------------------------------------
% - Related work  -----------------------------------------------------------
% ---------------------------------------------------------------------------
\section{Related work}
% JAX-based envs
\paragraph{JAX-based environments.}
The number of JAX-based reimplementations of common environments is in a bullish trend.
\citet{brax2021github} provide a fully differentiable physics engine for robotics, including MJX, a reimplementation of MujoCo \citep{todorov2012mujoco}.
\citet{gymnax2022github} reimplements several gym \citep{brockman2016openai} environments, including classic control, Bsuite \citep{osband2020bsuite}, and MinAtar \citep{young2019minatar},

\citet{koyamada2023pgx} reimplement many board games, including backgammon, chess, shogi, and go.
\citet{lu2023structured} provides JAX implementations of POPGym \citep{morad2023popgym}, which contains partially-observed RL environments.
\citet{matthews2024craftax} reimplement Crafter \citep{hafner2021crafter}.
\citet{bonnet2024jumanji} provides JAX implementations of combinatorial problems frequently encountered in industry, including bin packing, capacitated vehicle routing problem, PacMan, Sokoban, Snake, 2048, Sudoku, and many others.
\citet{rutherford2023jaxmarl} reimplement a set of multi-agent environments, including a MiniGrid-inspired implementation of the Overcooked benchmark.

Yet, none of these works proposes a reimplementation of Minigrid.
\citet{weng2022envpool} is the only one providing a single environment of the suite, \textit{Empty}, but it is only one of the many, most commonly used environments of the suite, and arguably the simplest one.

% Minigrid-like envs
\paragraph{Batched MiniGrid$-$like environments.}
Two works stand out for they aim to partially reimplement MiniGrid.
\citet{jiang2023minimax} present \texttt{AMaze}, a fully batched implementation of a partially observable maze environment, with MiniGrid$-$like sprites and observations.
However, like \citet{weng2022envpool}, the work does not reimplement the full MiniGrid suite.
\citet{nikulin2023xlandMiniGrid} proposes \texttt{XLand-MiniGrid}, a suite of grid-world environments for meta \gls{rl}.
Like \citep{jiang2023minimax}, \texttt{XLand-MiniGrid} reproduces Minigrid$-$like observations but focuses on designing a set of composable rules that can be used to generate a wide range of environments, rather than reimplementing the original Minigrid suite.

To conclude, MiniGrid is a fundamental tool for \gls{drl} experiments, at the base of a high number of studies, as we highlighted in Section~\ref{sec:intro}.
%  \citep{zhang2020bebold,zha2021rank,mavor2022stay,jiang2021prioritized,zhao2021consciousness,paischer2022history,flet2021adversarially,guan2022leveraging,parisi2021interesting,mu2022improving}.
It is easy to use, easy to extend, and provides a wide range of environments of scalable complexity that are easy to inspect for a clearer understanding of an algorithm dynamics, pitfalls, and strengths.

Nevertheless, none of the works above provides a full, batched reimplementation of Minigrid in JAX that mirrors the original suite in terms of environments, observations, state transitions, and rewards.
Instead, we propose a full JAX-based reimplementation of the MiniGrid suite that can be used as a drop-in replacement for the original environments.
% ---------------------------------------------------------------------------

% ---------------------------------------------------------------------------
% - Background  -------------------------------------------------------------
% ---------------------------------------------------------------------------
% \section{Notation}
% RL terms, state space, rewards, etc..
% Vector signatures, f32[], i32[], etc..
% ---------------------------------------------------------------------------

% ---------------------------------------------------------------------------
% - Methods  ----------------------------------------------------------------
% ---------------------------------------------------------------------------
\section{NAVIX: design philosophy and principles}
In this section we describe:
\begin{enumerate*}[label=\textit{\textbf{(\roman*)}}]
    \item the design philosophy and pattern of NAVIX in Section~\ref{sec:design}, and
    \item the design principles at its foundation in Sections~\ref{sec:matching} and \ref{sec:jittable}.
\end{enumerate*}

In particular, in Section~\ref{sec:jittable}, we highlight \emph{why} a JAX port of MiniGrid is not trivial.
Among others, the obstacles to transform a stateless program, where a function is allowed to change elements that are not an input of the function, to a stateful one, where the outputs of functions depend solely on the inputs; and the restrictions in the use of \texttt{for} loops and control flow primitives, such as \texttt{if} statements.\footnote{See \url{https://jax.readthedocs.io/en/latest/notebooks/Common_Gotchas_in_JAX.html}.}
\looseness=-1

\subsection{Design pattern}
\label{sec:design}
\begin{figure}[!t]
    \includegraphics[width=\textwidth]{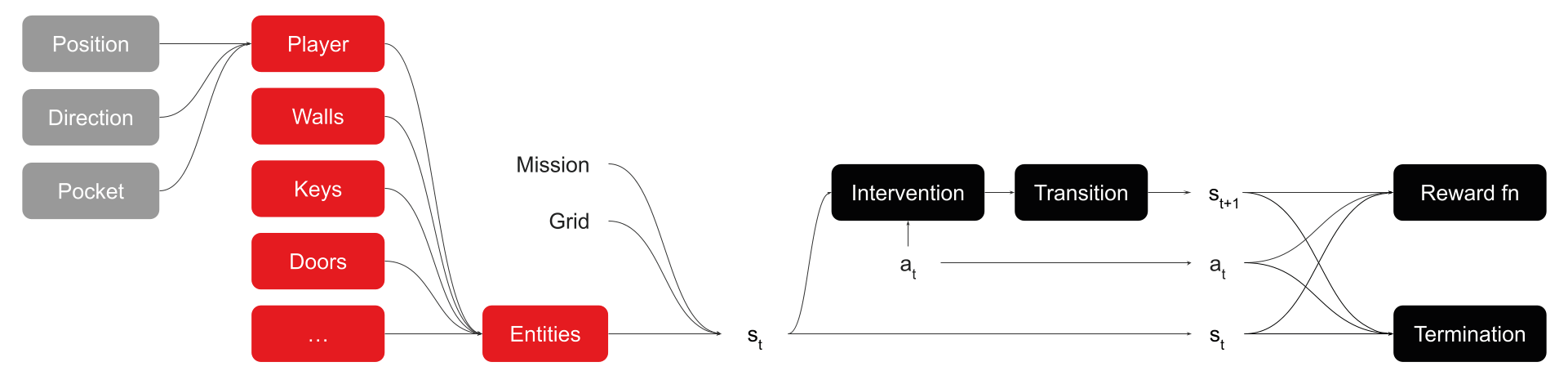}
    \captionof{figure}{
        Information flow of the \gls{ecsm} in NAVIX.
        Entities (\textit{Player, Walls, Keys, Doors, \ldots}) are composed of components (\textit{Position, Direction, Pocket}), which hold the data of the entity.
        Systems (\textit{Intervention, Transition, Rewards, Terminations}) are functions that operate on the collective state of all entities and components.
    }
    \label{fig:ecsm}
    \vspace{1em}
    % \end{figure}

    % \begin{table}
    \begin{tabular}{llll}
        \toprule
        \textbf{Component}    & \textbf{Property}    & \textbf{Shape}       & \textbf{Description}                                      \\
        \midrule
        \texttt{Positionable} & \texttt{Position}    & \texttt{f32[2]}      & \textit{Coordinates of the entity on the grid.          } \\
        \texttt{Directional}  & \texttt{Direction}   & \texttt{i32[]}       & \textit{Direction of the entity.                        } \\
        \texttt{HasColour}    & \texttt{Colour}      & \texttt{u8[]}        & \textit{Colour of the entity.                           } \\
        \texttt{Stochastic}   & \texttt{Probability} & \texttt{f32[]}       & \textit{Probability that the entity emits an event.     } \\
        \texttt{Openable}     & \texttt{State}       & \texttt{bool[]}      & \textit{State of the entity, e.g., open or closed.      } \\
        \texttt{Pickable}     & \texttt{Id}          & \texttt{i32[]}       & \textit{Id of the entity that the agent can pick up.    } \\
        \texttt{HasTag}       & \texttt{Tag}         & \texttt{i32[]}       & \textit{Categorical value identifying the entity class. } \\
        \texttt{HasSprite}    & \texttt{Sprite}      & \texttt{u8[32x32x3]} & \textit{Sprite of the entity in RGB format.             } \\
        \texttt{Holder}       & \texttt{Pocket}      & \texttt{i32[]}       & \textit{Id of the entity that the agent holds.          } \\
        \bottomrule
    \end{tabular}
    \vspace{1em}
    \captionof{table}{
        List of \textbf{Components} in NAVIX. Each component provides a property (or a set of).
        These properties hold the data that can be accessed and manipulated by the systems (see Table~\ref{tab:systems}) to provide observations, rewards, and state transitions.
    }
    \label{tab:components}

    \vspace{1em}
    \begin{tabular}{lll}
        \toprule
        \textbf{Entity} & \textbf{Components}              & \textbf{Description}                                                \\
        \midrule
        \texttt{Wall}   & \texttt{[HasColour]}             & \textit{An entity that blocks the agent's movement.}                \\
        \texttt{Player} & \texttt{[Directional, Holder]}   & \textit{An entity that can interact with the environment.}          \\
        \texttt{Goal}   & \texttt{[HasColour, Stochastic]} & \textit{An entity that the agent can to reach to receive a reward.} \\
        \texttt{Key}    & \texttt{[Pickable, HasColour]}   & \textit{An entity that can be picked up. Can open doors.}           \\
        \texttt{Door}   & \texttt{[Openable, HasColour]}   & \textit{An entity that can be opened and closed by the agent.}      \\
        \texttt{Lava}   & \texttt{[]}                      & \textit{An entity that the agent has to avoid.}                     \\
        \texttt{Ball}   & \texttt{[HasColour, Stochastic]} & \textit{An entity that the agent can push.}                         \\
        \texttt{Box}    & \texttt{[HasColour, Holder]}     & \textit{An entity that the agent can push.}                         \\
        \bottomrule
    \end{tabular}
    \vspace{1em}
    \captionof{table}{
        List of \textbf{Entities} in NAVIX, together with the components that characterise them.
        By default, all entities already possess \texttt{Positionable}, \texttt{HasTah}, and \texttt{HasSprite} components, in addition to those reported in the table.
    }
    \label{tab:entities}

    \vspace{1em}
    \begin{tabular}{lll}
        \toprule
        \textbf{System}       & \textbf{Function}                                                                  & \textbf{Description}                                            \\
        \midrule
        \texttt{Intervention} & $I: \mathcal{S} \times \mathcal{A} \rightarrow \mathcal{S}$                        & \textit{Updates the state according to the agent's actions.}    \\
        \texttt{Transition}   & $P: \mathcal{S} \times \mathcal{A} \rightarrow \mathcal{S}$                      & \textit{Updates the state according to the \gls{mdp} dynamics.} \\
        \texttt{Observation}  & $O: \mathcal{S} \rightarrow \mathcal{O}$                                           & \textit{The observation kernel;}                                \\
        \texttt{Reward}       & $R: \mathcal{S} \times \mathcal{A} \times \mathcal{S} \rightarrow \mathbb{R}$      & \textit{The Markovian reward function.}                         \\
        \texttt{Termination}  & $\gamma: \mathcal{S} \times \mathcal{A} \times \mathcal{S} \rightarrow \mathbb{B}$ & \textit{The termination function.}                              \\
        \bottomrule
    \end{tabular}
    \vspace{1em}
    \captionof{table}{
        List of \textbf{Systems} in NAVIX.
        A state $s \in \mathcal{S}$ is a tuple containing: the set of entities, the static grid layout, and the mission of the agent.
    }
    \label{tab:systems}
\end{figure}

% ECSM
NAVIX is broadly inspired by the \gls{ecsm}, a design pattern widely used in video game development.
In an \gls{ecsm}, entities -- the \textit{objects} on the grid in our case -- are composed of components -- the \textit{properties} of the object.
Each property holds data about the entity, which can then be used to process the game state.
For example, an entity \texttt{Player} is composed of components \texttt{Positionable}, \texttt{Holder}, \texttt{Directional}, each of which injects properties into the entity: the \texttt{Positionable} component injects the \texttt{Position} property, holding the coordinates of the entity (e.g., a player, a door, a key) on the grid, the \texttt{Holder} component injects the \texttt{Pocket} property, holding the id of the entity that the agent holds, and so on.
A full list of components and their properties is provided in Table~\ref{tab:components}.
This compositional layout allows to easily generate the wide range of combinations of tasks that MiniGrid offers, and to easily extend the suite with new environments.

Entities are then processed by \textit{systems}, which are functions that operate on the collective state of all entities and components.
For example, the \textit{decision} system may update the state of the entities according to the actions taken by a player.
The \textit{transition} system may update the state of the entities according to the \gls{mdp} state transitions.
The \textit{observation} system generates the observations that the agents receive, and the \textit{reward} system computes the rewards that the agents receive, and so on.
We provide a full list of implemented systems in Appendix~\ref{app:systems}.

To develop a better intuition of what these elements are and how they interact, Figure~\ref{fig:ecsm} shows the information flow of the \gls{ecsm} in NAVIX.
% Table~\ref{tab:components} lists the components of NAVIX, Table~\ref{tab:entities} the entities, and Table~\ref{tab:systems} enumerate the systems.

\subsection{Design principles}
On this background, two principles are at the foundation of NAVIX, and the key aspects that characterise it:
\begin{enumerate*}[label=\textit{\textbf{(\roman*)}}]
    \item NAVIX aims to exactly match MiniGrid (Section~\ref{sec:matching}), working as a drop-in replacement for the original environments, and;
    \item every environment is fully jittabile and differentiable (Section~\ref{sec:jittable}), to exploit the full set of features that JAX offers.
    % \item systems provide mathematical transparency (Section~\ref{sec:math-transparency});
    % \item NAVIX offers an easy entry point for new users (Section~\ref{sec:easy}).
\end{enumerate*}

\subsubsection{NAVIX matches MiniGrid}
\label{sec:matching}
NAVIX matches the original MiniGrid suite in terms of environments, observations, state transitions, rewards, and actions.
We include the most commonly used environments of the suite (see Table~\ref{tab:envs}), and provide a set of baselines for the implemented environments in see Section~\ref{sec:experiments} and Table~\ref{tab:envs}, Appendix~\ref{app:envs}.

Formally, a NAVIX environment is a tuple $\mathcal{M} = (h, w, T, \mathcal{O}, \mathcal{A}, \mathcal{R}, d, O, R, \gamma, P)$.
Here, $h$ and $w$ are the height and width of the grid, $T$ is the number of timesteps before timeout; $\mathcal{O}$ is the observation space, $\mathcal{A}$ is the action space, $\mathcal{R}$ is the reward space; $\gamma$ is the discount factor.
$O$ is the observation function, $R$ is the reward function, $d$ is the termination function, and $P$ is the transition function.

By default, one key difference between NAVIX and MiniGrid is that the latter uses a non-Markovian reward function.
In fact, MiniGrid dispenses a reward of $0$ everywhere, except at task completion, where it is inversely proportional to the number of steps taken by the agent to reach the goal:
\begin{align}
    r_t = R(s_t, a, s_{t+1}) - 0.9 * (t + 1) / T,
\end{align}
Here $R$ is the reward function, $s_t$ is the state at time $t$, $a$ is the action taken at time $t$, $s_{t+1}$ is the state at time $t+1$, and $T$ is the number of timesteps before timeout.
Notice the dependency on the number of steps $t$, which makes the reward non-Markovian.

The use of a non-Markovian reward function is not a mild assumption as most \gls{rl} algorithms assume Markov rewards.
This might call into question the validity of the historical results obtained with MiniGrid, and the generalisation of the results to other environments.
For this reason, we depart from the original MiniGrid reward function and use a Markovian reward function, instead, which is $0$ everywhere, and $1$ at task completion.

% \subsection{Easy to use}
% \label{sec:easy}
% NAVIX provides a simple interface that allows the agent to interact with the environment in a stateful way.

\subsubsection{Stateful and fully jittable}
\label{sec:jittable}
While we aim to match MiniGrid in terms of environments, observations, state transitions, rewards, and actions, the API of NAVIX is different, as it must align with JAX requirements for the environment to be fully jittable.
In fact, NAVIX environments can be compiled into XLA and run on any JAX-supported accelerator, including GPUs and TPUs.
This includes both simply jitting the \texttt{step} function, and jitting the entire training sequence \citep{lu2022discovered}, assuming that the agent is also implemented in JAX.
XLA compilation increases the throughput of experiments massively, allowing for the training of thousands of agents in parallel on a single accelerator, compared to a few that are possible with traditional CPU-based environments.
We show the scalability of NAVIX in Section~\ref{sec:experiments}.

% \paragraph{The timestep.}
For environments to be fully jittable, the computation must be stateful.
For this reason, we need to define an environment \textit{state-object}: the \textit{timestep}.
The timestep is a tuple $(t, o_t, a_t, r_{t+1}, \gamma_{t+1}, s_t, i_{t+1})$, where $t$ is the current time -- the number of steps elapsed from the last reset -- $o_t$ is the observation at time $t$, $a_t$ is the action taken after $o_t$, $r_{t+1}$ is the reward received after $a_t$, $\gamma_{t+1}$ is the termination signal after $a_t$, $s_t$ is the true state of the environment at time $t$, and $i_{t+1}$ is the info dictionary, useful to store accumulations, such as returns.

This structure is necessary to guarantee the same return schema for both the \texttt{step} and the \texttt{reset} methods, and allows the environment to autoreset, and avoid conditional statements in the agent code, which would prevent the environment from being fully jittable.

% \paragraph{Interaction protocol.}
At the beginning of the episode, the agent samples a starting state from the starting distribution $P_0: \mathcal{S} \rightarrow \mathcal{S}$ using the \texttt{reset(key)} method, where \texttt{key} is a key for a stateful random number generator.
Since there is no action and reward at the beginning of the episode, we pad with $-1$ and $0$, respectively.
Given an action $a_t$, the agent can interact with the environment by calling the \texttt{step(timestep, action, key)} method.
The agent then receives a new state of the environment (a new timestep) and can continue to interact as needed.
Code~\ref{code:interaction} shows an example of how to interact with a jitted NAVIX environment.
More examples are provided at \url{https://epignatelli.com/navix/}.

\begin{snippet}{}
    \begin{minted}{python}
    import navix as nx

    # init a NAVIX environment
    env = nx.make("Navix-KeyCorridorS6R3-v0")

    # sample a starting state
    timestep = env.reset(key)
    for _ in range(1000):
        # sample a random key
        key, subkey = jax.random.split(key)
        # sample a random action
        action = jax.random.randint(subkey, (1,), 0, 4)
        # interact with the environment
        timestep = jax.jit(env.step)(timestep, action)  # autoresets when done
    \end{minted}
    \caption{
        Example code to interact with a jitted NAVIX environment.
    }
    \label{code:interaction}
\end{snippet}

Notice that the syntax is similar to the original MiniGrid, including the environment \textit{id}, which simply replaces ``MiniGrid'' with ``Navix''.
The only differences are in the use of an explicit random key for the stateful random number generator, and the fact that the \texttt{step} method also takes the current timestep as input, to guarantee the statefulness of the computation.

The schema in Code~\ref{code:interaction} is an effective template for any kind of agent implementation, including non JAX-jittable agents.
However, while this already improves the speed of environment interactions compared to MiniGrid, as shown in Section~\ref{sec:experiments:speed}, the real speed-up comes jitting the whole iteration loop.
In Appendix~\ref{app:jitting} we provide additional reusable patterns that are useful in daily \gls{rl} research, such as how to jit the training loop, how to parallelise the training of multiple agents, and how to run hyperparameter search in batch mode.

In addition, in Appendix~\ref{app:extend} we provide a guide on how to extend NAVIX, including new environments, new observations, new rewards, and new termination functions.
This is a fundamental aspect to reflect the flexibility of the original MiniGrid suite, which is easy to extend and modify.

% ----------------------------------------------------------------------------------------------------
% --- Experiments ------------------------------------------------------------------------------------
% ----------------------------------------------------------------------------------------------------
\section{Experiments}
\label{sec:experiments}
This section aims to show the advantages of NAVIX compared to the original MiniGrid implementation, and provides the community with a set of baselines for all environments.
It does the former by comparing the two suites, for all environments, both in terms of speed improvements and throughput.
For the latter, we train a set of baselines for all environments, and provide a scoreboard that stores the results for all environments.
All experiments are run on a single Nvidia A100 80Gb, and Intel(R) Xeon(R) Silver 4310 CPU @ 2.10GHz and 128Gb of RAM.

\subsection{Speed}
\label{sec:experiments:speed}
We first benchmark the raw speed improvements of NAVIX compared to the original Minigrid implementation, in the most common settings.
For each NAVIX environment and its MiniGrid equivalent, we run $1K$ steps with $8$ parallel environments, and measure the wall time of both.
Notice that this is the mere speed of the environment, and does not include the agent training.

\begin{figure}
    \centering
    \includegraphics[width=0.9\textwidth]{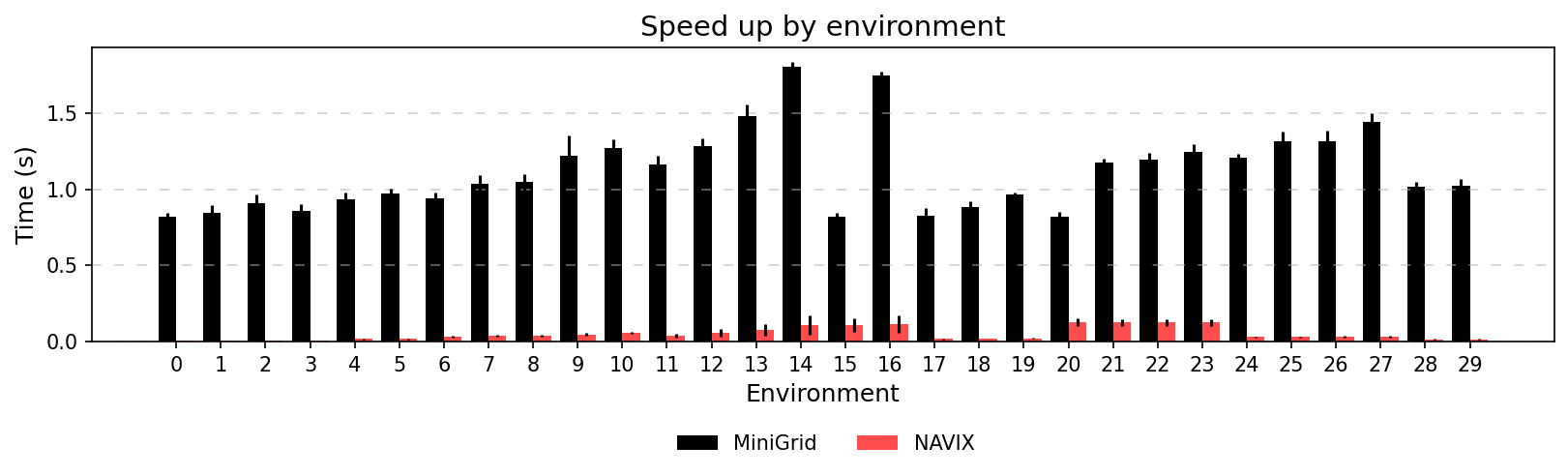}
    \caption{
        Speedup of NAVIX compared to the original Minigrid implementation, for the implemented environments.
        The identifiers on the x-axis correspond to the environments as reported in Table~\ref{tab:id-to-env}.
        Results are the average across 5 runs.
        Lines show $5$-$95$ percentile confidence intervals.
        Lower is better.
    }
    \label{fig:speedup-envs}

    \includegraphics[width=0.9\textwidth]{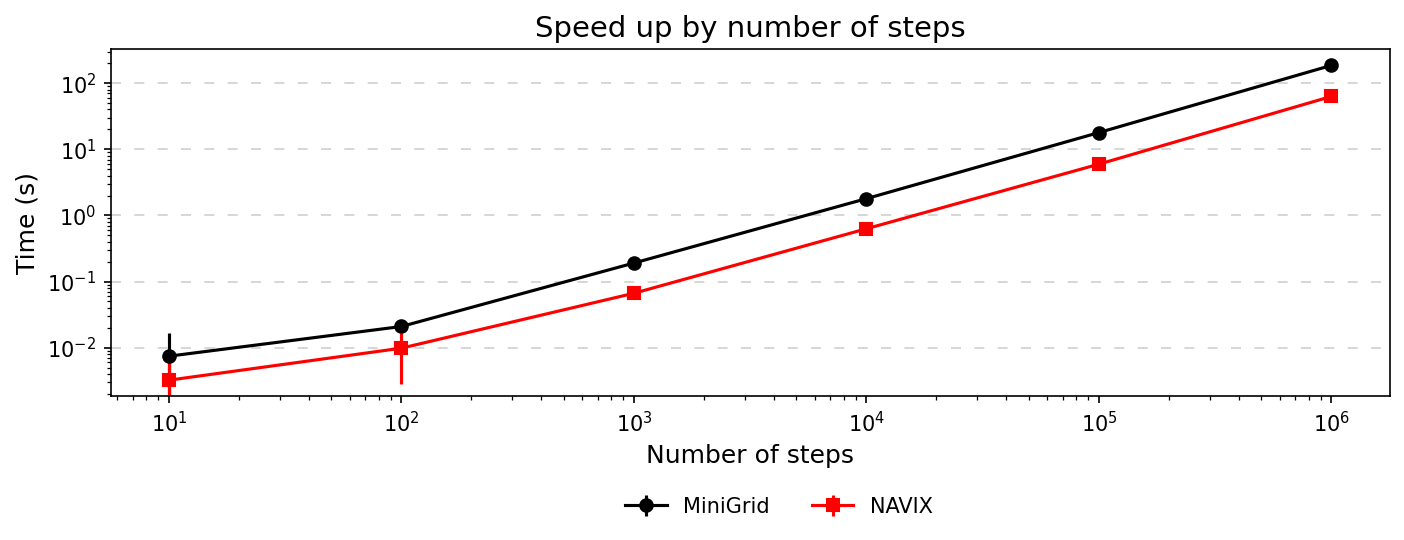}
    \caption{
        Variation of the speedup of NAVIX compared to the original Minigrid implementation according to different numbers of steps for the \texttt{MiniGrid-Empty-8x8-v0} environment.
        Lower is better. Error bars show $5$-$95$ percentile confidence intervals across 5 seeds.
    }
    \label{fig:speedup-steps}
\end{figure}

We show results in Figure~\ref{fig:speedup-envs} and observe that NAVIX is over $45\times$ faster than the original MiniGrid implementation on average.
These improvements are due to both the migration of the computation to the GPU via XLA, which optimises the computation graph for the specific accelerator, and the batching of the environments.
In Figure~\ref{fig:speedup-envs-single}, Appendix~\ref{app:envs} we ablate the batching, with no parallel environments, and show that the biggest contribution for the speedup is due to efficient batching.
% We note that only the \texttt{KeyCorridor} subset of environments shows a smaller speedup and sometimes even a slowdown, probably due to a bad implementation of the environment in NAVIX.

To better understand how the speedup varies with the number of training steps, and to make sure that the $1K$ steps used in the previous experiment are representative of the general trend, we measure the speed improvements for different lenghts of the training runs.
We run $1K$, $10K$, $100K$, and $1M$ steps for the \texttt{MiniGrid-Empty-8x8-v0} environment and its NAVIX equivalent, and measure the wall time of both.

Results in Figure~\ref{fig:speedup-steps} show that NAVIX is consistently faster than the original MiniGrid implementation, regardless of the number of steps.
Both MiniGrid show a linear increase in the wall time with the number of steps.

\subsection{Throughput}
\label{sec:experiments:throughput}
While NAVIX provides speed improvements compared to the original MiniGrid implementation, the real advantage comes from the ability to perform highly parallel training runs on a single accelerator.
In this experiment, we test how the computation scales with the number of environments.

We first test the limits of NAVIX by measuring the computation while varying the number of environments that run in parallel.
MiniGrid uses \texttt{gymnasium}, which parallelises the computation with \textit{Python}'s multiprocessing library.
NAVIX, instead, uses JAX's native \texttt{vmap}, which directly vectorises the computation.
We confront the results with the original MiniGrid implementation, using the \texttt{MiniGrid-Empty-8x8-v0} environment.

\begin{figure}
    \centering
    \includegraphics[width=0.9\textwidth]{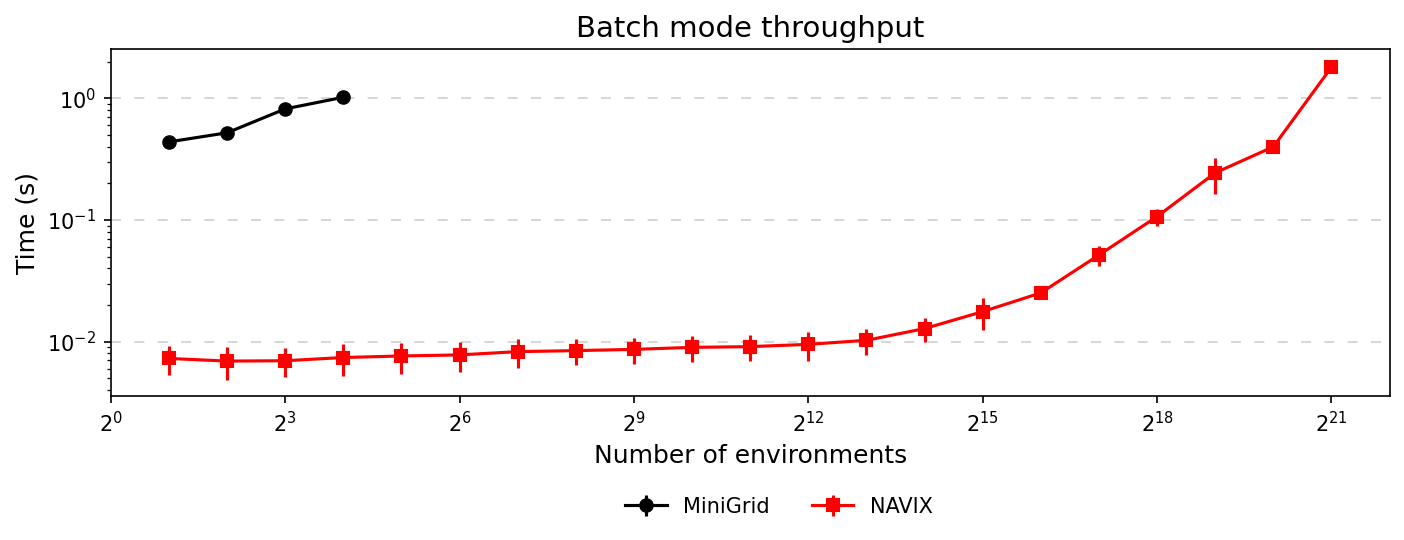}
    \caption{
        Wall time of $1K$ unrolls for both NAVIX and MiniGrid in batch mode.
    }
    \label{fig:throughput-envs}
\end{figure}

Results in Figure~\ref{fig:throughput-envs} show that the original MiniGrid implementation cannot scale beyond $16$ environments on 128GB of RAM, for which it takes around $1$s to complete $1K$ unrolls.
On the contrary, NAVIX can run up to $2^{21}$ (over $2M$) environments in parallel on the same hardware, with a wall time almost always below $1$s.
In short, NAVIX achieves a throughput over $10^5$ order of magnitude higher than the original MiniGrid implementation.

\begin{figure}
    \centering
    \includegraphics[width=0.9\textwidth]{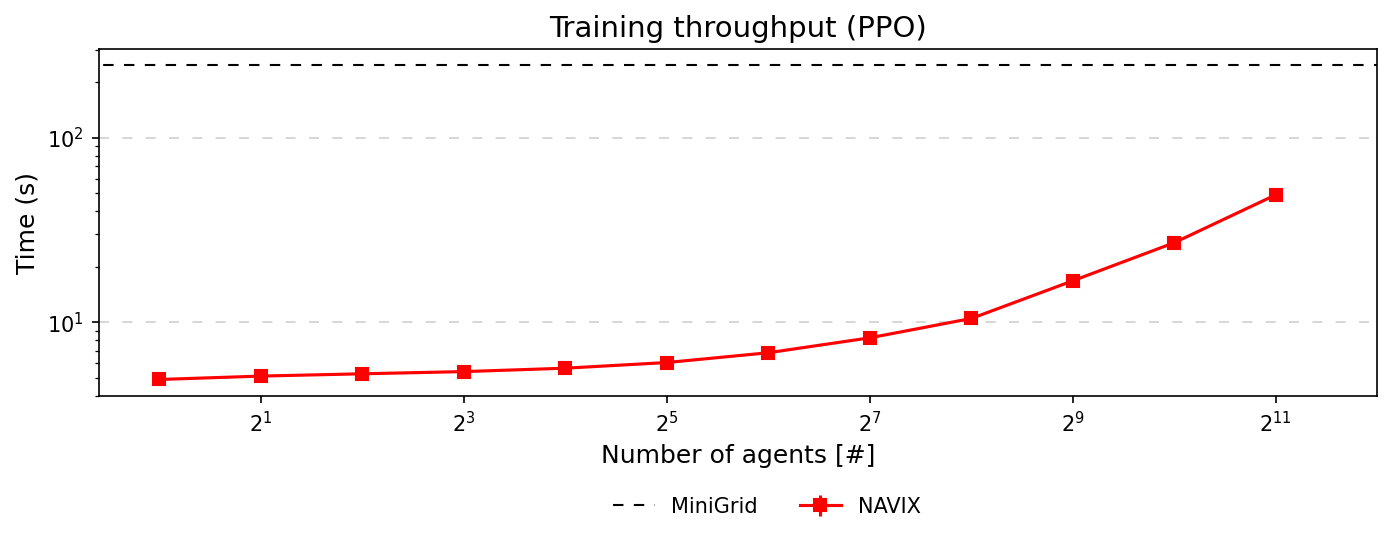}
    \caption{
        Computation costs with growing batch sizes.
        The agent is a PPO agent on a \textit{Navix-Empty-5x5} environment, run for $1M$ steps across 5 seeds.
        The effective number of environments is $16$ times the number of agents since each PPO agent works on $16$ environments.
    }
    \label{fig:throughput-ppo}
\end{figure}

Secondly, we simulate the very common operation of training many PPO agents, each with their own subset of 16 environments.
However, with NAVIX, we do this in parallel.
We set the \texttt{Empty-8x8-v0} environment, and train the agent for $1M$ steps.
Results are shown in Figure~\ref{fig:throughput-ppo}.

We observe that training $2048$ agents in NAVIX, for a total of $32\,768$ environments in parallel, takes less than $50$s, almost $5$ times faster than the original MiniGrid implementation, which takes around $240$s to train a \textit{single} PPO agent.
% In comparison, the original Minigrid implementation, with a CleanRL baseline takes around $240$s to train a \textit{single} PPO agent.
In other words, considering the performance at 2048 agents, NAVIX performs $2048 \times 1M / 49s = 668\,734\,693.88$ steps per second ($\sim 670$ Million steps/s) in batch mode, while the original Minigrid implementation performs $1M / 318.01 = 3\,144.65$ steps per second.
This is a speedup of over $200\,000\times$.

% This shows the scalability of NAVIX, and the ability to train thousands of agents in parallel on a single accelerator, compared to the few that are possible with traditional CPU-based environments.

\subsection{Baselines}
\label{sec:experiments:baselines}

\begin{figure}
    \centering
    \includegraphics[width=0.9\textwidth]{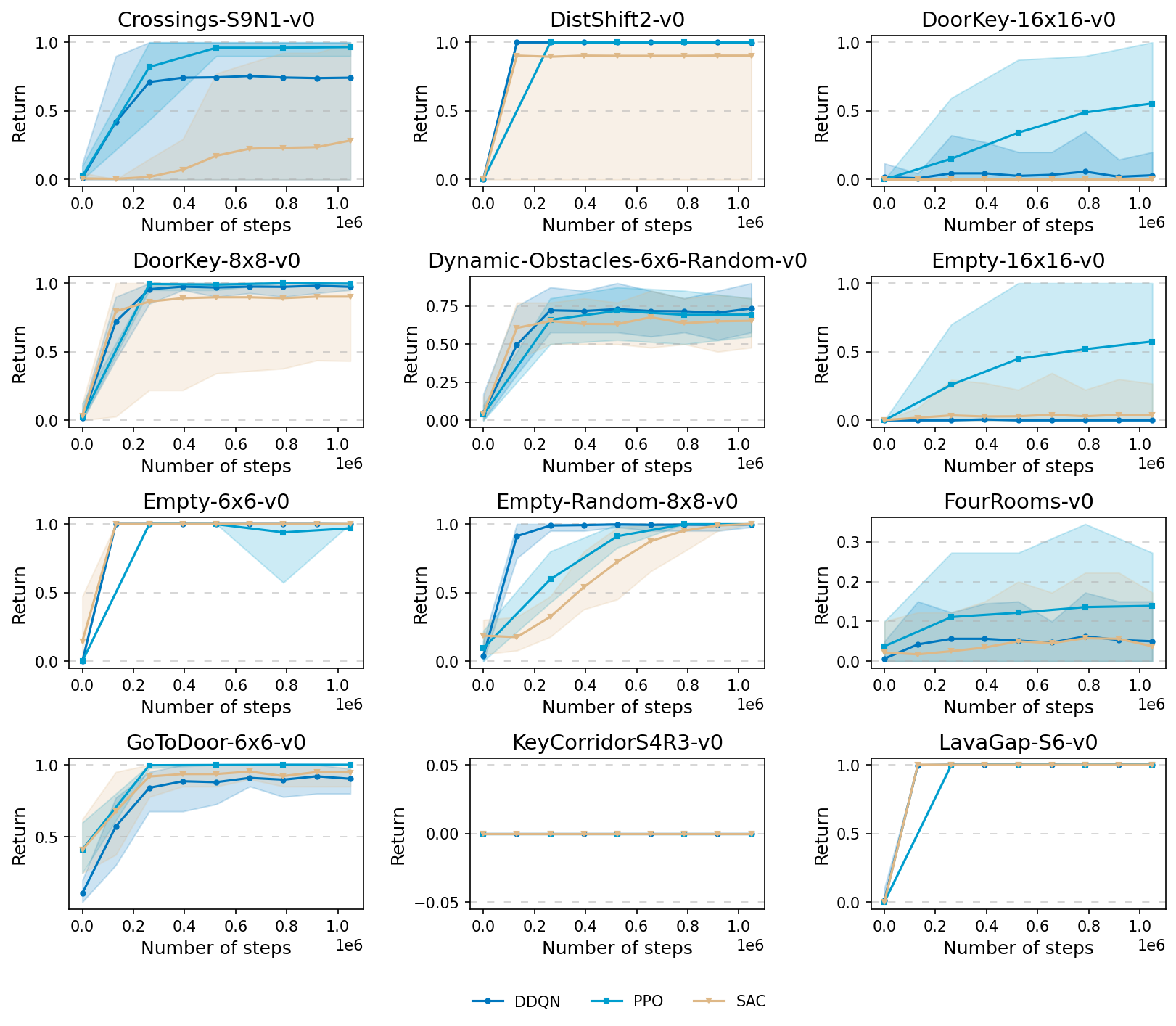}
    \caption{
        Episodic returns for a sample of NAVIX environments for DDQN, PPO and SAC baselines.
        Lines are average over 32 seeds, and shaded areas show $5$-$95$ percentile confidence intervals.
    }
    \label{fig:baselines}
\end{figure}
We provide additional baselines using the implementations of PPO \citep{schulman2017proximal}, Double DQN (DDQN) \citep{hasselt2016deep}, and Soft Actor Critic (SAC) \citep{haarnoja2018soft} in Rejax \citep{liesen2024rejax}.
We optimize hyperparameters (HP) for each algorithm and environment combination using 32 iterations of random search.
Each HP configuration is evaluated with 16 different initial seeds.
The HP configuration with the highest average final return is selected.
The specific hyperparameters we searched for are shown in Table~\ref{tab:fitted-hyperparameters}.

We run the baselines for $10M$ steps, across 32 seeds, with the tuned hyperparameters for the environments shown in Figure~\ref{fig:baselines}.
All algorithms use networks with two hidden layers of 64 units.
Instead of alternating between a single environment step and network update, DQN and SAC instead perform 128 parallel environment steps and 128 network updates, each with a new minibatch.
We found that this significantly improves the runtime while leaving the final performance unaffected.
\looseness=-1

% Finally, we provide a set of baselines for all environments that can be used as a term of reference for new algorithms.
% The baselines include our own implementation of PPO, available with NAVIX, and the ReJAX implementation of PPO, SAC \citep{haarnoja2018soft}, DDPG \citep{lillicrap2015continuous}, TD3 \citep{fujimoto2018addressing}, and DQN \citep{mnih2013playing}.
% We run the baselines for $10M$ steps, across 5 seeds, for all environments.
% Results are shown in Figure~\ref{fig:baselines}.
% We present the full list of results in Figure~\ref{fig:baselines-full}, Appendix~\ref{app:baselines}, we provide the code to run the baselines at \url{https://github.com/epignatelli/navix/baselines}, and an interactive visualisation, including the raw data at \url{https://epignatelli.com/navix/baselines/}.

% \begin{figure}
%     \centering
%     \includegraphics[width=\textwidth]{figures/baselines.png}
%     \caption{
%         Episodic returns for a sample of NAVIX environments for the PPO baseline.
%     }
%     \label{fig:baselines}
% \end{figure}

% \subsection{The effects of a non-Markovian reward function}
% \label{sec:experiments:non-markovian}
% ----------------------------------------------------------------------------------------------------

% ----------------------------------------------------------------------------------------------------
% --- Conclusions ------------------------------------------------------------------------------------
% ----------------------------------------------------------------------------------------------------
\section{Conclusions}
We introduced NAVIX, a reimplementation of the Minigrid environment suite in JAX that leverages JAX's intermediate language representation to migrate the computation to different accelerators, such as GPUs and TPUs.
We described the design philosophy, the design pattern, the organisation, and the components of NAVIX, highlighting the connections to the \gls{ecsm} design pattern, and the correspondence between the structure of its functions and the mathematical formalism of \gls{rl}.

We presented the environment interface, the list of available environments, and the scoreboard, storing state-of-the-art results that new algorithms can refer to avoid running also baselines, which are prone to errors and manipulations.
We showed the speed improvements of NAVIX compared to the original Minigrid implementation, and the scalability of NAVIX with respect to the number of agents that can be trained in parallel, or the number of environments that can be run in parallel.

Overall, NAVIX is over 1000x faster than the original Minigrid implementation, turning 1-week experiments into 15-minute ones.
With the current pace of the research in \gls{rl}, the ability to run fast experiments is crucial to keep up with the state-of-the-art, and to develop new, more efficient algorithms.
We hope that NAVIX will be a valuable tool for the \gls{rl} community, and that it will foster the development of new, more efficient algorithms, and the exploration of new research directions.
% ----------------------------------------------------------------------------------------------------

\bibliography{references}
% ----------------------------------------------------------------------------------------------------

% ----------------------------------------------------------------------------------------------------
% --- Appendix ---------------------------------------------------------------------------------------
% ----------------------------------------------------------------------------------------------------
\appendix
\section{Details on NAVIX systems}
\label{app:systems}
% Reward functions, transition functions, termination functions, observation functions, action-interventions.
Systems are \textit{functions} that operate on the collective state of all entities, defining the rules of the interactions between them.
In designing NAVIX, we aimed to maintain a bijective relationship between the systems and their respective mathematical formalism in \gls{rl}.
This makes it easier to translate the mathematical formalism into code, and vice versa, connecting the implementation to the theory.
NAVIX includes the following systems:
\begin{enumerate*}
    \item \texttt{Intervention}: a function that updates the state of the entities according to the actions taken by the agents.
    \item \texttt{Transition}: a function that updates the state of the entities according to the \gls{mdp} state transitions.
    \item \texttt{Observation}: a function that generates the observations that the agents receive.
    \item \texttt{Reward}: a function that computes the rewards that the agents receive.
    \item \texttt{Termination}: a function that determines if the episode is terminated.
\end{enumerate*}
We now describe the systems formally.

The intervention is a function $I: \mathcal{S} \times \mathcal{A} \rightarrow \mathcal{S}$ that updates the state of the entities according to the actions taken by the agents.
This corresponds to the canonical decision in an \gls{mdp}.

The transition is a function $\mu: \mathcal{S} \times \mathcal{A} \rightarrow \mathcal{S}$ that updates the state of the entities according to the \gls{mdp} state transitions.
This corresponds to the canonical state transition kernel in an \gls{mdp}.

The observation is a function $O: \mathcal{S} \rightarrow \mathcal{O}$ that generates the observations that the agents receive.
NAVIX includes multiple observation functions, each generating a different type of observation, for example, a first-person view, a top-down view, or a third-person view, both in symbolic and pixel format.
We provide both full and partial observations, allowing to cast a NAVIX environment both as an \gls{mdp} or as a \gls{pomdp}, depending on the needs of the algorithm.
This follows the design of the original MiniGrid suite.

The reward is a function $R: \mathcal{S} \times \mathcal{A} \rightarrow \mathbb{R}$ that computes the rewards that the agents receive.
Likewise, the termination is a function $\gamma: \mathcal{S} \rightarrow \{0, 1\}$ that determines if the episode is terminated.
We include both the reward and the termination functions necessary to reproduce all MiniGrid environments.
Both these systems rely on the concept of \textit{events}, representing a goal to achieve.
An \textit{event} is itself an entity signalling that a particular state of the environment has been reached.
For example, it can indicate that the agent has reached a particular cell, has picked up a particular object, or that the agent performed a certain action in a particular state.

We provide a summary of the implemented systems in NAVIX in Tables~\ref{tab:observations},~\ref{tab:rewards}, and~\ref{tab:terminations} for the observation, reward, and termination systems, respectively.

\begin{table}[h]
    \begin{tabular}{p{0.33\linewidth} p{0.33\linewidth} p{0.33\linewidth}}
        \toprule
        \textbf{Observation function}       & \textbf{Shape}                 & \textbf{Description}                                                     \\
        \midrule
        \texttt{symbolic}                   & \texttt{i32[H, W, 3]}          & \textit{The canonical \textit{grid encoding} observation from MiniGrid.} \\
        \texttt{symbolic\_first\_person}    & \texttt{i32[R, R, 3]}          & \textit{A first-person view of the environment in symbolic format.}      \\
        \texttt{rgb}                        & \texttt{u8[32 * H, 32 * W, 3]} & \textit{A fully visible image of the environment in RGB format.}         \\
        \texttt{rgb\_first\_person}         & \texttt{u8[32 * R, 32 * R, 3]} & \textit{A first-person view of the environment in RGB format.}           \\
        \texttt{categorical}                & \texttt{i32[H, W]}             & \textit{A grid of entities ID in the environment.}                       \\
        \texttt{categorical\_first\_person} & \texttt{i32[R, R]}             & \textit{A first-person view of the grid of entities ID.}
    \end{tabular}
    \vspace{1em}
    \caption{Implemented observation functions in NAVIX.}
    \label{tab:observations}

    \begin{tabular}{p{0.33\textwidth} p{0.67\textwidth}}
        \toprule
        \textbf{Reward function}   & \textbf{Description}                                                                                                         \\
        \midrule
        \texttt{on\_goal\_reached} & \textit{$+1$ when a Goal entity and a Player entity have the same position}                                                  \\
        \texttt{on\_lava\_fall}    & \textit{$-1$ when a Lava entity and a Player entoty have the same position}                                                  \\
        \texttt{on\_door\_done}    & \textit{$+1$ when the \texttt{done} action is performed in front of a door with the colour specific in the \texttt{mission}} \\
        \texttt{free}              & $0$ everywhere                                                                                                               \\
        \texttt{action\_cost}      & \textit{$-cost_a$ at every action taken, except \texttt{done}}                                                               \\
        \texttt{time\_cost}        & \textit{$-cost_t$ at every step}
    \end{tabular}
    \vspace{1em}
    \caption{Implemented reward functions in NAVIX.}
    \label{tab:rewards}

    \begin{tabular}{p{0.33\linewidth} p{0.67\linewidth}}
        \toprule
        \textbf{Termination function} & \textbf{Description}                                                                                                               \\
        \midrule
        \texttt{on\_goal\_reached}    & \textit{Terminates when a Goal entity and a Player entity have the same position}                                                  \\
        \texttt{on\_lava\_fall}       & \textit{Terminates when a Lava entity and a Player entity have the same position}                                                  \\
        \texttt{on\_door\_done}       & \textit{Terminates when the \texttt{done} action is performed in front of a door with the colour specific in the \texttt{mission}} \\
        \texttt{free}                 & $0$ everywhere                                                                                                                     \\
    \end{tabular}
    \vspace{1em}
    \caption{Implemented termination functions in NAVIX.}
    \label{tab:terminations}
\end{table}

\section{Reusable patterns}
\label{app:jitting}
Here we provide some useful patterns that users can reuse as-they-are or modify to suit their needs.
In particular, we show how to jit the full interaction loop of a NAVIX environment in Code~\ref{code:jit-loop}, and how to run multiple seeds in parallel in Code~\ref{code:parallel-seeds}.
Further examples, including how to jit a whole training loop with a JAX-based agent, and how to automate hyperparameter search, are available in the NAVIX documentation at \url{https://epignatelli.com/navix/examples/getting_started.html}.

\subsection{Jitting full interaction loops}
\begin{snippet}{}
    \begin{minted}{python}
    import navix as nx

    # init a NAVIX environment
    env = nx.make("Navix-KeyCorridorS6R3-v0")

    # sample a starting state
    timestep = env.reset(key)

    # jitting the step function
    step_env = jax.jit(env.step)
    
    # unroll the environment for 1000 steps
    timestep, _ = jax.lax.scan(
        lambda timestep, _: (unroll(timestep, i % 6), ()),
        timestep,
        (timestep, jnp.arange(1000))
    )

    \end{minted}
    \caption{Example code to jit a \texttt{Navix-Empty-5x5-v0} environment.}
    \label{code:jit-loop}
\end{snippet}

\subsection{Running multiple seeds in parallel}
\begin{snippet}{}
    \begin{minted}{python}
    import navix as nx

    env = nx.make("Navix-KeyCorridorS6R3-v0")
    
    # define the run function
    def run(key):
        def step(state, action):
            timestep, key = state
            key, subkey = jax.random.split(key)
            action = jax.random.randint(subkey, (), 0, env.action_space.n)
            return (env.step(timestep, action), key), ()


        # unroll the environment for 1000 steps
        timestep = env.reset(key)
        timestep, _ = jax.lax.scan(
            step,
            timestep,
            ((timestep, key) jnp.arange(1000)),
        )
        return timestep
    
    seeds = jax.random.split(jax.random.PRNGKey(0), 1000)
    batched_end_steps = jax.jit(jax.vmap(run))(seeds)

    \end{minted}
    \caption{Example code to jit a \texttt{Navix-Empty-5x5-v0} environment.}
    \label{code:parallel-seeds}
\end{snippet}

\section{Customising NAVIX environments}
\label{app:custom}
NAVIX is designed to be highly customisable, allowing users to create new environments by combining existing entities and systems.
In this section, we provide examples of how to customise NAVIX environments by using different \textit{systems}.

For example, to create a new environment where the agent has to reach a goal while avoiding lava, we can combine the \texttt{Goal} and \texttt{Lava} entities with the \texttt{Reward} system:
\begin{snippet}{}
    \begin{minted}{python}
    import navix a nx

    reward_fn = nx.rewards.compose(
        nx.rewards.on_goal_reached(),
        nx.rewards.on_lava_fall()
    )

    env = nx.make(
        "Navix-Empty-5x5-v0",
        reward_fn=reward_fn)
    \end{minted}
    \caption{
        Example code to create a \texttt{Navix-Empty-5x5-v0} environment with a custom reward function.
        See Table~\ref{tab:rewards} for a list of implemented reward functions.
    }
\end{snippet}

Alternatively, to use a different observation function, we can use the \texttt{Observation} system:
\begin{snippet}{}
    \begin{minted}{python}
    import navix as nx

    env = nx.make(
        "Navix-Empty-5x5-v0",
        observation_fn=nx.observations.rgb())
    \end{minted}
    \caption{
        Example code to create a \texttt{Navix-Empty-5x5-v0} environment with a custom observation function.
        See Table~\ref{tab:observations} for a list of implemented observation functions.
    }
\end{snippet}

Finally, to terminate the environment, for example, only when the agent reaches the goal, but not when it falls into the lava, we can use the \texttt{Termination} system:
\begin{snippet}{}
    \begin{minted}{python}
    import navix as nx

    env = nx.make(
        "Navix-Empty-5x5-v0",
        termination_fn=nx.terminations.on_goal_reached())
    \end{minted}
    \caption{
        Example code to create a \texttt{Navix-Empty-5x5-v0} environment with a custom termination function.
        See Table~\ref{tab:terminations} for a list of implemented termination functions.
    }
\end{snippet}

These examples can be extended to create more complex environments by combining different systems for the same environment configuration.

\section{Extending NAVIX environments}
\label{app:extend}
NAVIX is designed to be easily extensible.
Users can create new entities, components, systems, and full environments by implementing the necessary functions.
In this section, we provide \textbf{templates} to extend NAVIX environments.
In particular, Code~\ref{code:custom-env} shows how to create a custom environment, Code~\ref{code:custom-component} shows how to create a custom component, Code~\ref{code:custom-entity} shows how to create a custom entity, and Code~\ref{code:custom-systems} shows how to create custom systems.

\begin{snippet}{}
    \begin{minted}{python}
import jax, navix as nx

class CustomEnv(nx.Environment):
    def _reset(self, key: jax.Array) -> nx.Timestep:
        """Reset the environment."""
        # create your grid, place your entities, define your mission
        return timestep


nx.registry.register_env(
    "CustomEnv",
    lambda *args, **kwargs: CustomEnv.create(
        observation_fn=nx.observations.symbolic(),
        reward_fn=nx.rewards.on_goal_reached(),
        termination_fn=nx.terminations.on_goal_reached(),
    )
)
\end{minted}
    \caption{
        Example code to extend NAVIX by creating a custom environment. The \texttt{\_reset} function allows to generate a custom starting state, after which the environment will evolve according to the usual systems: intervention, transition, reward and termination functions.
        Notice that it is convenient to use the environment constructor \texttt{create} to automatically set non-orthogonal properties (e.g. observation space and observation function).
    }
    \label{code:custom-env}
\end{snippet}

\begin{snippet}{}
    \begin{minted}{python}
import jax, navix as nx

class CustomComponent(nx.Componnet):
    """My custom component."""

    custom_property: jax.Array = nx.components.field(shape=())
    \end{minted}
    \caption{
        Example code to extend NAVIX by creating a custom component.
        Notice that the property must have a type annotation and specify a shape.
    }
    \label{code:custom-component}
\end{snippet}

\begin{snippet}{}
    \begin{minted}{python}
import jax, navix as nx

class CustomEntity(nx.Entity, CustomComponent):
    """My custom entity."""

    @property
    def walkable(self) -> jax.Array:
        return jnp.broadcast_to(jnp.asarray(False), self.shape)

    @property
    def transparent(self) -> jax.Array:
        return jnp.broadcast_to(jnp.asarray(False), self.shape)

    @property
    def sprite(self) -> jax.Array:
        sprite = # the address of your sprite, e.g., SPRITES_REGISTRY[Entities.WALL]
        return jnp.broadcast_to(sprite[None], (*self.shape, *sprite.shape))

    @property
    def tag(self) -> jax.Array:
        entity_id = # the id of your entity, e.g., EntityIds.WALL
        return jnp.broadcast_to(entity_id, self.shape)
    \end{minted}
    \caption{
        Example code to extend NAVIX by creating a custom entity.
        Notice that four properties must be implemented: \texttt{walkable}, \texttt{transparent}, \texttt{sprite}, and \texttt{tag}.
    }
    \label{code:custom-entity}
\end{snippet}

\begin{snippet}{}
    \begin{minted}{python}
import jax, navix as nx

def my_reward_function(state: nx.State, action: nx.Action, new_state: nx.State) -> jax.Array:
    """My custom reward function."""
    # do stuff
    return reward  # f32[]

def my_termination_function(state: nx.State, action: nx.Action, new_state: nx.State) -> jax.Array:
    """My custom termination function."""
    # do stuff
    return termination  # bool[]

def my_observation_function(state: nx.State) -> jax.Array:
    """My custom observation function."""
    # do stuff
    return observation  # f32[]

def my_intervention_function(state: nx.State, action: nx.Action) -> nx.State:
    """My custom intervention function."""
    # do stuff
    return new_state  # State

def my_transition_function(state: nx.State) -> nx.State:
    """My custom transition function."""
    # do stuff
    return new_state  # State
\end{minted}
    \caption{
        Example code to extend NAVIX by creating custom systems.
    }
    \label{code:custom-systems}
\end{snippet}
\FloatBarrier

\section{Additional Tables}
\label{app:envs}

\begin{figure}
    \centering
    \includegraphics[width=0.9\textwidth]{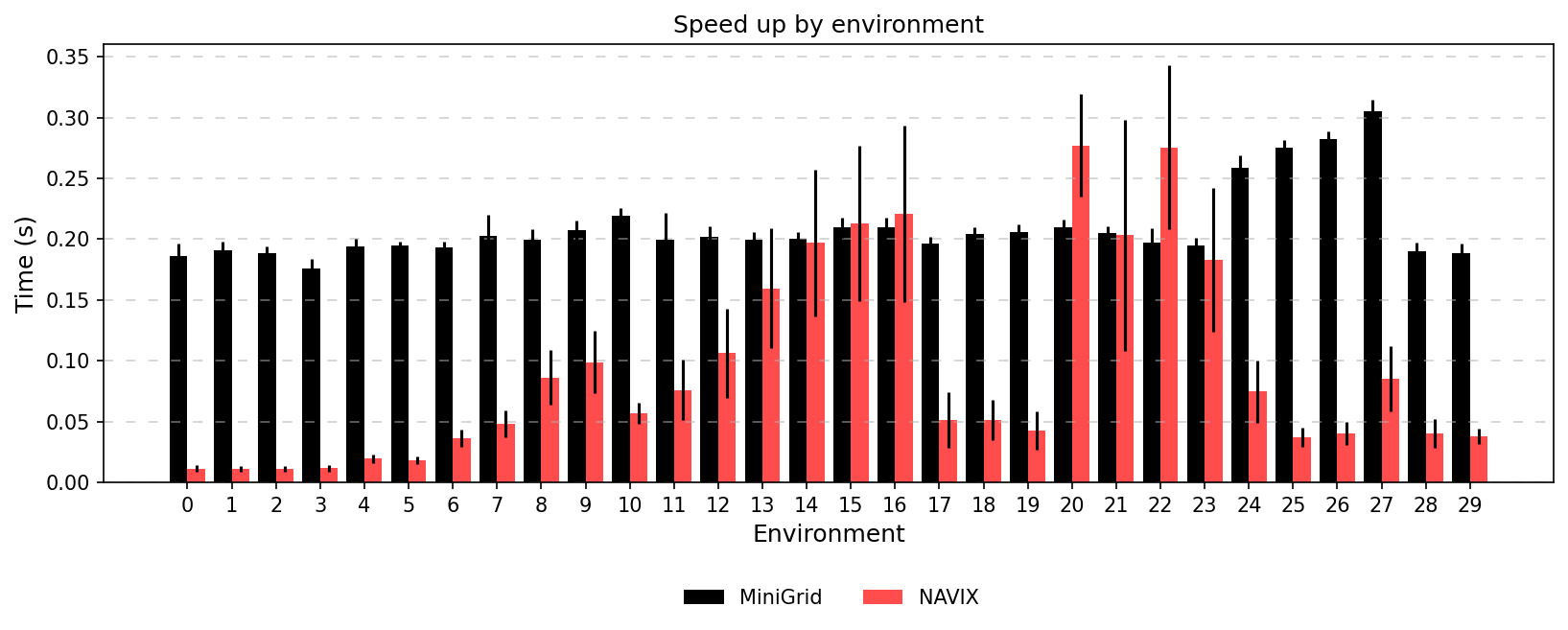}
    \caption{
        \textbf{Ablation.} Speedup of NAVIX compared to the original Minigrid implementation without batching.
        The identifiers on the x-axis correspond to the environments as reported in Table~\ref{tab:id-to-env}.
        Lower is better.
    }
    \label{fig:speedup-envs-single}
\end{figure}

\begin{table}[H]
    \centering
    \begin{tabular}{ll}
        \toprule
        \textbf{X tick} & \textbf{Env id}               \\
        \midrule
        0               & Navix-Empty-5x5-v0            \\
        1               & Navix-Empty-6x6-v0            \\
        2               & Navix-Empty-8x8-v0            \\
        3               & Navix-Empty-16x16-v0          \\
        4               & Navix-Empty-Random-5x5        \\
        5               & Navix-Empty-Random-6x6        \\
        6               & Navix-DoorKey-5x5-v0          \\
        7               & Navix-DoorKey-6x6-v0          \\
        8               & Navix-DoorKey-8x8-v0          \\
        9               & Navix-DoorKey-16x16-v0        \\
        10              & Navix-FourRooms-v0            \\
        11              & Navix-KeyCorridorS3R1-v0      \\
        12              & Navix-KeyCorridorS3R2-v0      \\
        13              & Navix-KeyCorridorS3R3-v0      \\
        14              & Navix-KeyCorridorS4R3-v0      \\
        15              & Navix-KeyCorridorS5R3-v0      \\
        16              & Navix-KeyCorridorS6R3-v0      \\
        17              & Navix-LavaGapS5-v0            \\
        18              & Navix-LavaGapS6-v0            \\
        19              & Navix-LavaGapS7-v0            \\
        20              & Navix-SimpleCrossingS9N1-v0   \\
        21              & Navix-SimpleCrossingS9N2-v0   \\
        22              & Navix-SimpleCrossingS9N3-v0   \\
        23              & Navix-SimpleCrossingS11N5-v0  \\
        24              & Navix-Dynamic-Obstacles-5x5   \\
        25              & Navix-Dynamic-Obstacles-6x6   \\
        26              & Navix-Dynamic-Obstacles-8x8   \\
        27              & Navix-Dynamic-Obstacles-16x16 \\
        28              & Navix-DistShift1-v0           \\
        29              & Navix-DistShift2-v0           \\
        % 30              & Navix-GoToDoor-5x5-v0         \\
        % 31              & Navix-GoToDoor-6x6-v0         \\
        % 32              & Navix-GoToDoor-8x8-v0         \\
        \bottomrule
    \end{tabular}
    \vspace{1em}
    \caption{
        Correspondence between the x-ticks in Figure~\ref{fig:speedup-envs} and the environment ids.
    }
    \label{tab:id-to-env}
\end{table}

Table of environments available in NAVIX.
\begin{table}[H]
    % \rowcolors{2}{gray!10}{white}
    \centering
    \begin{tabular}{lllll}
        \toprule
        % \rowcolor{gray!25}
        \textbf{Env-id}                      & \textbf{Class}    & \textbf{Height} & \textbf{Width} & \textbf{Reward} \\
        \midrule
        Navix-Empty-5x5-v0                   & Empty             & 5               & 5              & $R_1$           \\
        Navix-Empty-6x6-v0                   & Empty             & 6               & 5              & $R_1$           \\
        Navix-Empty-8x8-v0                   & Empty             & 8               & 8              & $R_1$           \\
        Navix-Empty-16x16-v0                 & Empty             & 16              & 16             & $R_1$           \\
        Navix-Empty-Random-5x5               & Empty             & 5               & 5              & $R_1$           \\
        Navix-Empty-Random-6x6               & Empty             & 6               & 6              & $R_1$           \\
        Navix-Empty-Random-8x8               & Empty             & 8               & 8              & $R_1$           \\
        Navix-Empty-Random-16x16             & Empty             & 16              & 16             & $R_1$           \\
        \toprule
        Navix-DoorKey-5x5-v0                 & DoorKey           & 5               & 5              & $R_1$           \\
        Navix-DoorKey-6x6-v0                 & DoorKey           & 6               & 6              & $R_1$           \\
        Navix-DoorKey-8x8-v0                 & DoorKey           & 8               & 8              & $R_1$           \\
        Navix-DoorKey-16x16-v0               & DoorKey           & 16              & 16             & $R_1$           \\
        Navix-DoorKey-Random-5x5             & DoorKey           & 5               & 5              & $R_1$           \\
        Navix-DoorKey-Random-6x6             & DoorKey           & 6               & 6              & $R_1$           \\
        Navix-DoorKey-Random-8x8             & DoorKey           & 8               & 8              & $R_1$           \\
        Navix-DoorKey-Random-16x16           & DoorKey           & 16              & 16             & $R_1$           \\
        \toprule
        Navix-FourRooms-v0                   & FourRooms         & 17              & 17             & $R_1$           \\
        \toprule
        Navix-KeyCorridorS3R1-v0             & KeyCorridor       & 3               & 7              & $R_1$           \\
        Navix-KeyCorridorS3R2-v0             & KeyCorridor       & 5               & 7              & $R_1$           \\
        Navix-KeyCorridorS3R3-v0             & KeyCorridor       & 7               & 7              & $R_1$           \\
        Navix-KeyCorridorS4R3-v0             & KeyCorridor       & 10              & 10             & $R_1$           \\
        Navix-KeyCorridorS5R3-v0             & KeyCorridor       & 13              & 13             & $R_1$           \\
        Navix-KeyCorridorS6R3-v0             & KeyCorridor       & 16              & 16             & $R_1$           \\
        \toprule
        Navix-LavaGap-S5-v0                  & LavaGap           & 5               & 5              & $R_2$           \\
        Navix-LavaGap-S6-v0                  & LavaGap           & 6               & 6              & $R_2$           \\
        Navix-LavaGap-S7-v0                  & LavaGap           & 7               & 7              & $R_2$           \\
        \toprule
        Navix-Crossings-S9N1-v0              & Crossings         & 9               & 9              & $R_2$           \\
        Navix-Crossings-S9N2-v0              & Crossings         & 9               & 9              & $R_2$           \\
        Navix-Crossings-S9N3-v0              & Crossings         & 9               & 9              & $R_2$           \\
        Navix-Crossings-S11N5-v0             & Crossings         & 11              & 11             & $R_2$           \\
        \toprule
        N        Navix-Dynamic-Obstacles-5x5 & Dynamic-Obstacles & 5               & 5              & $R_3$           \\
        Navix-Dynamic-Obstacles-5x5          & Dynamic-Obstacles & 5               & 5              & $R_3$           \\
        Navix-Dynamic-Obstacles-6x6          & Dynamic-Obstacles & 6               & 6              & $R_3$           \\
        Navix-Dynamic-Obstacles-6x6          & Dynamic-Obstacles & 6               & 6              & $R_3$           \\
        Navix-Dynamic-Obstacles-8x8          & Dynamic-Obstacles & 8               & 8              & $R_3$           \\
        Navix-Dynamic-Obstacles-16x16        & Dynamic-Obstacles & 16              & 16             & $R_3$           \\
        \toprule
        Navix-DistShift1-v0                  & DistShift         & 6               & 6              & $R_2$           \\
        Navix-DistShift2-v0                  & DistShift         & 8               & 8              & $R_2$           \\
        \toprule
        Navix-GoToDoor-5x5-v0                & GoToDoor          & 5               & 5              & $R_1$           \\
        Navix-GoToDoor-6x6-v0                & GoToDoor          & 6               & 6              & $R_1$           \\
        Navix-GoToDoor-8x8-v0                & GoToDoor          & 8               & 8              & $R_1$           \\
        % \bottomrule
    \end{tabular}
    \vspace{1em}
    \caption{
        List of environments available in NAVIX.
        \textit{Env-id} denotes the id to instantiate the environment.
        Here, $R_1$ is the reward function for goal achievement -- $1$ when the agent is on the green square, and $0$ otherwise.
        $R_2$ is the reward function for goal achievement and lava avoidance -- $1$ when the agent is on the green square, $-1$ when the agent is on the lava square, and $0$ otherwise.
        $R_3$ is the reward function for goal achievement and dynamic obstacles avoidance -- $1$ when the agent is on the green square, $-1$ when the agent is hit by a flying object, and $0$ otherwise.
        All environments terminate when the reward is not $0$, for example, on goal achievement, or on lava collision.
    }
    \label{tab:envs}
\end{table}
\FloatBarrier

% \section{Baselines}

\section{Additional details on baselines}
\begin{table}[H]
    \centering
    \begin{tabular}{ll}
        \toprule
        Algorithm & Fitted hyperparameters                                                                       \\
        \midrule
        PPO       & \#envs, \#steps, \#epochs, \#minibatches, discount factor, $\lambda$ (GAE), grad. norm clip, \\
                  & norm. obs., activation function                                                              \\
        DQN       & batch size, target network update freq., discount factor, exploration fraction,              \\
                  & final $\epsilon$, grad. norm clip, norm. obs., activation function                           \\
        SAC       & batch size, discount factor, $\tau$ (Polyak update), target entropy ratio, norm. obs.,       \\
                  & activation function                                                                          \\
        \bottomrule
    \end{tabular}
    \vspace{1em}
    \caption{
        Fitted hyperparameters for PPO, DQN, and SAC.
    }
    \label{tab:fitted-hyperparameters}
\end{table}

Details on each hyperparameter set, for each environment and each algorithm are available at \url{https://github.com/keraJLi/rejax/tree/main/configs}.

% \subsection{}
% \begin{table}[h]
%     \centering
%     \begin{tabular}{ll}

%     \end{tabular}
%     \caption{Fitted hyperparameters for DDQN.}
% \end{table}

% \begin{table}[h]
%     \centering
%     \begin{tabular}{ll}

%     \end{tabular}
%     \caption{Fitted hyperparameters for PPO.}
% \end{table}

% \begin{table}[h]
%     \centering
%     \begin{tabular}{ll}

%     \end{tabular}
%     \caption{Fitted hyperparameters for SAC.}
% \end{table}

\end{document}